\documentclass[10pt,twocolumn,letterpaper]{article}

\usepackage{cvpr}

\usepackage{lineno}
\usepackage{comment}
\newcommand\blfootnote[1]{%
  \begingroup
  \renewcommand\thefootnote{}\footnote{#1}%
  \addtocounter{footnote}{-1}%
  \endgroup
}
\usepackage[accsupp]{axessibility}

\definecolor{cvprblue}{rgb}{0.21,0.49,0.74}
\usepackage[pagebackref,breaklinks,colorlinks,allcolors=cvprblue]{hyperref}

\title{MEGA: Masked Generative Autoencoder for Human Mesh Recovery}

\author{
    Guénolé Fiche\textsuperscript{1,2$^\ast$} \hspace{0.5cm} 
    Simon Leglaive\textsuperscript{1} \hspace{0.5cm}  
    Xavier Alameda-Pineda\textsuperscript{3} \hspace{0.5cm} 
    Francesc Moreno-Noguer\textsuperscript{4$^{\dagger}$} \vspace{0.1cm} \\
    \textsuperscript{1}CentraleSupélec (IETR UMR CNRS 6164) \hspace{0.5cm}
    \textsuperscript{2}Naver Labs Europe \\
    \textsuperscript{3}Inria, Univ. Grenoble Alpes, CNRS, LJK \hspace{0.5cm}
    \textsuperscript{4}Amazon
}

\begin{document}
\maketitle
\blfootnote{$^\ast$ Work done at CentraleSupélec before joining Naver Labs Europe.}
\blfootnote{$^\dagger$ Work done at IRI (CSIC-UPC) before joining Amazon.}

\begin{abstract}
Human Mesh Recovery (HMR) from a single RGB image is a highly ambiguous problem, as an infinite set of 3D interpretations can explain the 2D observation equally well. Nevertheless, most HMR methods overlook this issue and make a single prediction without accounting for this ambiguity. A few approaches generate a distribution of human meshes, enabling the sampling of multiple predictions; however, none of them is competitive with the latest single-output model when making a single prediction. 
This work proposes a new approach based on masked generative modeling. By tokenizing the human pose and shape, we formulate the HMR task as generating a sequence of discrete tokens conditioned on an input image. We introduce MEGA, a MaskEd Generative Autoencoder trained to recover human meshes from images and partial human mesh token sequences. Given an image, our flexible generation scheme allows us to predict a single human mesh in deterministic mode or to generate multiple human meshes in stochastic mode. Experiments on in-the-wild benchmarks show that MEGA achieves state-of-the-art performance in deterministic and stochastic modes, outperforming single-output and multi-output approaches. See the project page at \href{https://g-fiche.github.io/research-pages/mega/}{https://g-fiche.github.io/research-pages/mega/}.
\end{abstract}    
\section{Introduction}
\label{sec:intro}

Perceiving humans from images is a long-standing problem in computer vision, with applications in diverse fields such as sports~\cite{swim, AIcoach} or e-commerce~\cite{liu2020comparing, zheng2019virtually}. Many approaches rely on statistical body models like SMPL~\cite{loper2015smpl} for representing humans. Earlier human mesh recovery (HMR) methods recovered the SMPL pose and shape parameters from 2D cues using optimization-based techniques~\cite{bogo2016keep, lassner2017unite}. However, these optimization procedures require good initialization, are time-consuming, and often converge to suboptimal minima. With the advancement of deep learning and the availability of datasets of images with 3D human pose and shape annotations, most approaches have shifted to a regression-based paradigm. Early approaches used neural architectures based on convolutional neural networks~(CNNs) and multilayer perceptrons (MLPs)~\cite{hmrKanazawa17, li2022cliff}. Recent works have adopted  Transformers~\cite{vaswani2017attention} for extracting image features or making predictions~\cite{cho2022FastMETRO, goel2023humans}. Despite achieving unprecedented accuracy, state-of-the-art HMR models still have some weaknesses, such as producing unrealistic predictions, especially when dealing with occlusions. To address these issues, recent works have proposed tokenizing the human pose using vector quantized-variational autoencoders (VQ-VAEs)~\cite{fiche2023vq, dwivedi2024tokenhmr}. This approach uses a discrete representation of the human mesh, learned from large-scale motion capture datasets, to confine predictions to the space of anthropomorphic meshes using a dictionary of valid mesh tokens. This tokenization aligns well with Transformer-based architectures, which were initially designed for processing discrete data in natural language processing. Notably, VQ-HPS~\cite{fiche2023vq} reframed HMR as a classification task and achieved state-of-the-art results with large and small training datasets, demonstrating the great potential of human mesh tokenized representations in HMR.

\begin{figure*}[t!]
    \centering
    \includegraphics[width=\textwidth]{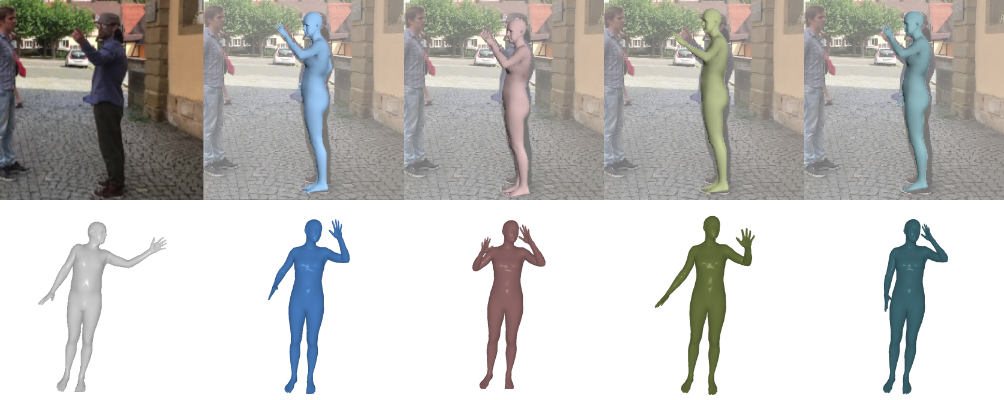}
    \caption{\textbf{Human mesh recovery from a single image is an ill-posed problem due to depth ambiguity.} Probabilistic approaches have aimed to address this by generating multiple predictions, but diversity often sacrifices accuracy. Introducing MEGA, our HMR model based on masked generative modeling achieves state-of-the-art performance on in-the-wild benchmarks in single- and multi-output settings. Given a single image, MEGA can make predictions that all look accurate given the 2D cues but correspond to diverse 3D interpretations.}
    \label{fig:teaser}
\end{figure*}

While significant progress has been made in HMR, a major issue remains unaddressed in most prior works: a 2D image, especially with occlusions, cannot provide sufficient information to estimate a 3D human mesh with certainty~\cite{moreno2012stochastic, simo2012single} (see ~\cref{fig:teaser}). This limitation causes single-output models to be biased toward most common poses and body shapes~\cite{corona2022learned}. To mitigate this problem, several works have proposed probabilistic approaches that generate multiple predictions from a single image~\cite{sengupta2021probabilistic, kolotouros2021probabilistic}. These approaches have used various families of generative models, ranging from conditional variational autoencoders (CVAEs)~\cite{sharma2019monocular} to diffusion models~\cite{cho2023generative, xu2024scorehypo}. However, this increase in diversity typically comes at the cost of accuracy~\cite{sengupta2023humaniflow}, and none of these multi-output methods are competitive with the latest single-output HMR models when making a single prediction.

In this work, we introduce MEGA, a multi-output HMR approach based on self-supervised learning and masked generative modeling of tokenized human meshes. MEGA relies on the Mesh-VQ-VAE of~\cite{fiche2023vq} to encode/decode a 3D human mesh to/from a set of discrete tokens. Our training process unfolds in two steps: (1) Firstly, akin to (vector quantized) masked autoencoders~\cite{he2022masked, bao2021beit, sadok2023vector, sadok2023avector}, we pre-train MEGA in a self-supervised manner to reconstruct human mesh tokens from partially visible inputs. This leverages an extensive motion capture dataset without needing paired image data, allowing MEGA to learn prior knowledge about 3D humans. (2) Subsequently, for HMR from RGB images, we train MEGA to predict randomly masked human mesh tokens conditioned on image feature embeddings. During inference, we begin with a fully masked sequence of tokens and generate a human mesh conditioned on an input image. We propose two distinct generation modes: (2.a) In deterministic mode, MEGA predicts all tokens in a single forward pass, ensuring speed and accuracy; (2.b) In stochastic mode, the generation process involves iteratively sampling human mesh tokens, enabling MEGA to produce multiple predictions from a single image.

We evaluate MEGA on in-the-wild HMR benchmarks, comparing it to single-output and probabilistic HMR methods. MEGA achieves state-of-the-art (SOTA) performance when predicting a human mesh in a single forward pass in deterministic mode. In stochastic mode, MEGA outperforms both deterministic and probabilistic methods with a single prediction and significantly enhances its performance with an increased number of samples. This mode generates diverse, realistic human meshes, allowing for the proposal of multiple plausible outputs given an image. This allows the user to choose the best-suited solution depending on the use case. For instance, animation designers may want to select the most visually appealing human mesh, while medical practitioners would require high precision and choose the solution that minimizes a reprojection error.

In summary, we make the following key contributions:
\begin{itemize}[left=10pt]
    \item We introduce MEGA, a masked generative autoencoder for human mesh recovery, pre-trained in a self-supervised manner on motion capture data.
    \item Our flexible inference procedure can operate in deterministic or stochastic modes, with or without image conditioning, achieving state-of-the-art results in all tested scenarios.
\end{itemize} 
\section{Related work}

\subsection{Human mesh recovery}

\noindent{\bf Single output HMR.} Since the release of the HMR~\cite{hmrKanazawa17} model, most approaches for recovering human meshes from images have been regression-based, using neural networks to make predictions directly from the image. These regression-based HMR methods can be categorized into parametric and non-parametric approaches. Parametric methods aim to recover the parameters of the SMPL model~\cite{sun2021monocular, Kocabas_PARE_2021, li2021hybrik, li2022cliff,goel2023humans, dwivedi2024tokenhmr, kolotouros2019learning, joo2021exemplar, pymaf2021, khirodkar2022occluded, li2023niki}. They typically produce realistic predictions; however, some works have argued that the SMPL model parameter space is not the most suitable for predicting human meshes~\cite{corona2022learned, kolotouros2019cmr, choi2020pose2mesh}, leading to the development of non-parametric approaches. Non-parametric methods predict the coordinates of 3D vertices without relying on a parametric model. While earlier approaches used graph convolutional neural network architectures inspired by the mesh topology~\cite{kolotouros2019cmr, lin2021-mesh-graphormer}, recent non-parametric models predominantly employ Transformers~\cite{cho2022FastMETRO, lin2021end-to-end, dou2023tore, fiche2023vq, ma20233d, kim2023sampling}. Although non-parametric methods yield accurate results, they can sometimes produce non-anthropomorphic meshes, particularly when training data is scarce~\cite{fiche2023vq}.

In this work, instead of predicting SMPL model parameters or 3D coordinates, we aim to recover sequences of token indices that can be decoded into a human mesh using the Mesh-VQ-VAE from~\cite{fiche2023vq}. Thus, the closest HMR method to ours is VQ-HPS~\cite{fiche2023vq}, which also works with the token representation of the Mesh-VQ-VAE. However, there are also important differences between MEGA and VQ-HPS. First, VQ-HPS follows a deterministic classification-based mapping between images and mesh tokens. This contrasts with MEGA's masked generative modeling approach, which allows multi-output predictions and unconditional human mesh generation (see \cref{app:rand_gen}). Second, the architecture of MEGA differs from that of VQ-HPS. The two models' encoders take different inputs: image patches for VQ-HPS and visible mesh tokens for MEGA. This key difference allows us to pre-train MEGA only with motion capture before conditioning the generation with images and to discard the encoder in the deterministic mode for faster inference. Ablation studies (see \cref{tab:quant_det}) show that this pre-training without any image conditioning significantly enhances the performance of MEGA. Another related HMR method is TokenHMR~\cite{dwivedi2024tokenhmr}, which is a regression method that uses human pose tokens as an intermediate representation.

\noindent{\bf Multi-output HMR.} Estimating a 3D human mesh from a single image is challenging due to the depth ambiguity, especially when the person is partially occluded. Several works have proposed making multiple predictions to account for the ill-posed nature of the problem. Earlier works employed compositional models~\cite{jahangiri2017generating} or mixture density networks~\cite{li2019generating}. More recent approaches rely on sophisticated probabilistic distributions~\cite{sengupta2021probabilistic, sengupta2021hierarchical} and generative models, such as CVAE~\cite{sharma2019monocular}, normalizing flows~\cite{kolotouros2021probabilistic, biggs2020, sengupta2023humaniflow}, and diffusion models~\cite{cho2023generative, xu2024scorehypo}. While these methods can predict diverse plausible solutions, they often face a trade-off between accuracy and diversity~\cite{sengupta2023humaniflow, xu2024scorehypo} and need to make a large number of predictions to be competitive with single-output methods.

MEGA is based on masked generative modeling to produce multiple predictions. Our experiments demonstrate that, while generating diverse samples, MEGA outperforms SOTA approaches even with a single prediction in stochastic mode and significantly improves as the number of samples increases.

\subsection{Self-supervised learning for HMR} 
Self-supervised learning (SSL) approaches can be categorized into two families: discriminative and generative~\cite{liu2021self, zhang2022survey, ozbulak2023know}. Many prior works used discriminative SSL approaches to train 3D human pose estimation models. Most of these methods exploit multi-view consistency constraints for supervision~\cite{kocabas2019self, wandt2021canonpose, roy2022triangulation, chen2019unsupervised}, while others use temporal consistency in videos~\cite{kundu2020self, schmidtke2022self, tung2017self} or images with different resolutions~\cite{xu20203d}. \cite{choi2022rethinking} explored the use of discriminative SSL for pre-training human mesh estimator backbones, demonstrating that 2D annotation-based pre-training leads to faster convergence and improved results. However,~\cite{armando2023cross} surpassed traditional feature extractors by employing generative SSL, using cross-view and cross-pose completion to train a Vision Transformer (ViT)~\cite{dosovitskiy2020image}. 

While prior works have demonstrated the importance of pre-training for the backbone of HMR models~\cite{Black_CVPR_2023, pang2022benchmarking}, we propose pre-training the generative model on human meshes to leverage extensive motion capture data. Different from prior works in this area~\cite{baradel2021leveraging, lin2024mpt}, our approach based on human mesh tokens does not need any rendering or reprojection, the masking happens in 3D. This pre-training provides us with an unconditional human mesh generative model (see \cref{app:rand_gen}), and ablation studies in \cref{tab:quant_det} show its important contribution to training the HMR model.

\subsection{Masked generative modeling} 
Masked modeling was introduced in BERT~\cite{kenton2019bert} for language modeling and extended to images with the masked autoencoder~\cite{he2022masked}. This technique trains a model to predict randomly masked tokens in a sequence based on visible tokens. Masked generative modeling builds on this by training a model to generate new samples, starting from a fully masked sequence and iteratively predicting a fixed number of tokens at each step~\cite{chang2022maskgit, chang2023muse, guo2023momask}.

In this work, we develop the first masked generative modeling approach for HMR. Using a tokenized representation of the human mesh is particularly well-suited for this task, as it allows for straightforward masking and replacement of mesh parts with mask tokens.
\section{MEGA}\label{sec:method}

\begin{figure*}[t!]
\centering
\includegraphics[width=\linewidth]{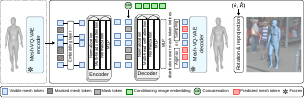}
\caption{MEGA is a masked generative model based on an encoder-decoder Transformer architecture. During the self-supervised pretraining stage, MEGA is trained to predict human mesh tokens from partially visible inputs using motion capture data without paired image data. During the supervised training stage for HMR, the model is trained to predict randomly masked human mesh tokens conditioned on image embeddings. For both training stages, only the cross-entropy loss is used on the predicted mesh tokens. At test time, in stochastic inference mode, we start from a fully masked sequence of tokens and iteratively sample human mesh tokens conditioned on input image embeddings. In deterministic inference mode, we predict all tokens in a single forward pass.
}
\label{fig:architecture_new}
\vspace{-2mm}
\end{figure*}

\subsection{Preliminary: Human mesh tokenization}\label{subsec:mesh-vq-vae}

MEGA relies on a tokenized representation of the human mesh. Specifically, we use the Mesh-VQ-VAE introduced in~\cite{fiche2023vq}, which is a VQ-VAE~\cite{van2017neural} with a fully convolutional mesh autoencoder architecture~\cite{zhou2020fully}. The Mesh-VQ-VAE tokenizes canonical human meshes following the SMPL topology, with zero translation and facing the camera. The input canonical mesh with vertices $V_c \in \mathbb{R}^{6890 \times 3}$ is encoded into a sequence of $N=54$ latent vectors, each of dimension $L=9$. As the architecture is fully convolutional, each latent vector encodes a specific part of the human body. Through vector quantization, each latent vector is replaced by a human mesh token index corresponding to an embedding vector of dimension $L$ in a codebook of size $S=512$. Thus, a human mesh is represented by a sequence of $N$ token indices in $\{1,...,S\}$, which can be decoded with the Mesh-VQ-VAE decoder to reconstruct the vertices $\hat{V}_c$. The Mesh-VQ-VAE is pre-trained on motion capture data and remains frozen during  MEGA's training. In this paper, we formulate HMR as the task of generating human mesh tokens conditioned on an input image.

\begin{figure*}
    \centering
    \includegraphics[width=0.9\textwidth]{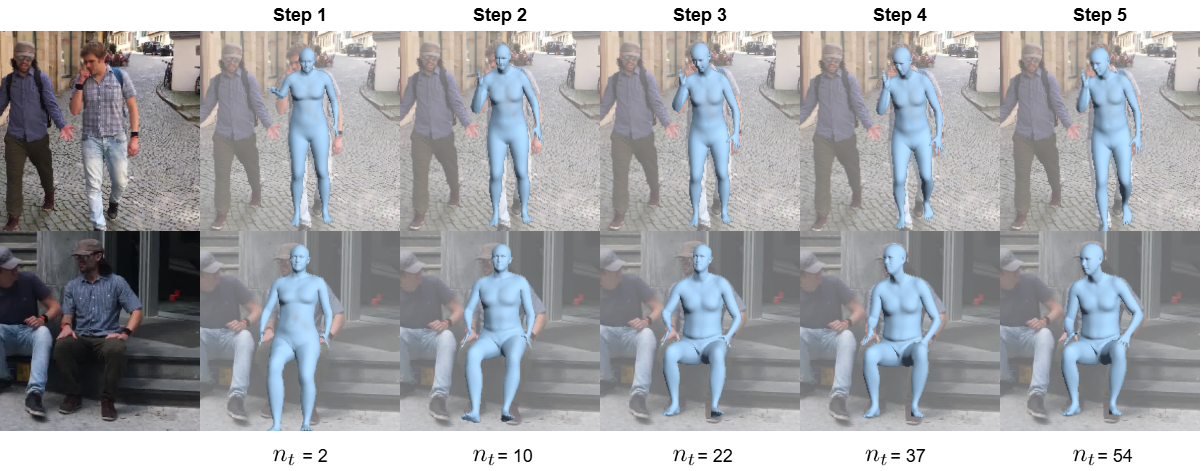}
    \caption{\textbf{Prediction process iterations.} We visualize the predictions for intermediate steps in stochastic mode. All masked tokens are replaced by the first token of the codebook, corresponding to index 0.}
    \label{fig:hmr_process}
\end{figure*}

\subsection{Model}\label{subsec:architecture}

\noindent{\bf Overall pipeline.} MEGA is a masked generative model based on an encoder-decoder Transformer architecture, illustrated in \cref{fig:architecture_new}. During training, the Mesh-VQ-VAE encoder converts a 3D human mesh into a sequence of $N$ tokens. These human mesh tokens are then randomly masked, leaving only $M < N$ visible tokens. An embedding layer converts the visible token indices into learned token embeddings, which are subsequently passed to an encoder with multi-head self-attention. The sequence of encoded token embeddings is completed with mask tokens and, during supervised training and inference stages only, it is also concatenated with a sequence of image embeddings (see next paragraph). The complete sequence of tokens is then processed by a decoder with multi-head self-attention, followed by an MLP that outputs a distribution over the $N$ human mesh token indices (in practice, it outputs the logits, which can be normalized using a softmax function). During training and in deterministic inference mode, the $N-M$ human mesh tokens that were originally masked are predicted by taking the argmax, and the Mesh-VQ-VAE decoder is used to reconstruct the canonical mesh $\hat{V}_c$. The cross-entropy loss is computed from the predicted mesh tokens for both the self-supervised and supervised training stages. At test time, the input sequence of mesh tokens is entirely masked (i.e., $M = 0$) such that the decoder is provided with $N$ mask tokens along with the conditioning image embeddings. In deterministic inference mode, the $N$ mesh tokens are predicted in a single forward pass. In stochastic inference mode, the human mesh tokens are predicted iteratively by sampling the predicted distribution over the mesh token indices and reintroducing a proportion of unmasked tokens at the encoder's input. Finally, an MLP (not represented in \cref{fig:architecture_new}) is trained to predict a global 6D rotation $\hat{R} \in \mathbb{R}^6$ and the perspective camera parameters $\hat{\pi} \in \mathbb{R}^3$ from image features. These can be used to orient the predicted canonical mesh and to reproject it on the input image.

\noindent{\bf Image embeddings extractor.} An image features extractor  computes image features $X \in \mathbb{R}^{WH \times C}$. For fair comparisons with SOTA methods, we rely on an HRNet~\cite{wang2020deep} backbone ($W=H=7$ and $C=720$) pre-trained on a 2D pose estimation task~\cite{lin2014microsoft}. Additionally, we provide results using a more powerful ViT~\cite{dosovitskiy2020image} backbone ($W=12$, $H=16$, $C=1280$). For the multi-output HMR experiment, we adhere to the common practice in the literature~\cite{cho2023generative, biggs2020, kolotouros2021probabilistic} and use a ResNet-50~\cite{he2016deep} backbone ($W=H=7$, $C=2048$). Prior to being fed to the MEGA decoder, the image features are linearly projected to the Transformer dimension, resulting in  $WH$ image embeddings of dimension $D=1024$.

\noindent{\bf Encoder.} The encoder of MEGA consists of $B_e=12$ blocks, akin to the Vision Transformer~\cite{dosovitskiy2020image}. Each block contains a multi-head self-attention and an MLP module, with layer normalization preceding and residual connections following every module (see \cref{fig:architecture_new}). To preserve positional information at the input of the encoder, learned position embeddings~\cite{gehring2017convolutional} are added to the human mesh token embeddings, and a \textit{cls} token~\cite{kenton2019bert} is concatenated with the input sequence.

\noindent{\bf Decoder.} The decoder of MEGA consists of $B_d=4$ blocks identical to the encoder blocks. During the self-supervised pre-training on human meshes (see \cref{subsec:train}), the decoder receives only the sequence of encoded human mesh token embeddings, completed with mask tokens (a single trainable embedding vector repeated at each masked token position), in line with the masked autoencoder strategy~\cite{he2022masked}. During the supervised training stage and at inference, this input sequence is concatenated with the sequence of image embeddings. Position embeddings~\cite{vaswani2017attention,gehring2017convolutional} are added at the input of the decoder.

\noindent{\bf Rotation and camera prediction.} We use the  previously mentioned image features $X \in \mathbb{R}^{WH \times C}$ to  predict  the global 6D rotation $\hat{R} \in \mathbb{R}^6$ and the perspective camera parameters $\hat{\pi} \in \mathbb{R}^3$. The $WH$ image features are averaged to yield a single vector of dimension $C$, which is subsequently passed through an MLP with 2 hidden layers. The output of this MLP is then linearly mapped to the rotation and camera parameters.

\subsection{Training strategy}\label{subsec:train}

\noindent{\bf Self-supervised pre-training.} MEGA is pre-trained in a self-supervised manner on tokenized human meshes using a strategy similar to vector quantized masked autoencoders~\cite{he2022masked, bao2021beit, sadok2023vector, sadok2023avector}. The pre-training task involves reconstructing randomly masked human mesh tokens from a set of visible tokens. A variable masking rate is used such that $M = \lfloor N \cos(\frac{\pi \tau}{2}) \rfloor$ with $\tau$ uniformly sampled from $[0,1[$. The variable masking rate is critical for allowing MEGA to generate meshes iteratively in stochastic mode, as each step of the generation process involves predicting all tokens given a variable number of visible tokens (see \cref{subsec:inference}). The sole loss used for pre-training is a cross-entropy loss computed from the reconstruction of the masked tokens. At this stage of training, MEGA can be used for the unconditioned generation of random human meshes, as shown in~\cref{app:rand_gen}

\noindent{\bf Masked generative modeling for HMR.} To train MEGA to predict tokenized canonical meshes from images, we extend the pre-training strategy by conditioning the decoder with the image embeddings (see \cref{subsec:architecture} and \cref{fig:architecture_new}). The human mesh tokens follow the same masking rate schedule, while the image embeddings remain fully visible.  The only supervision for predicting canonical meshes is a cross-entropy loss, as in the pre-training stage. Compared to prior works in HMR that rely on multiple losses (on 3D keypoints, the 2D reprojection, and the SMPL parameters), this approach is straightforward and does not require hyperparameters tuning. Note that the performance of Mesh-VQ-VAE~\cite{fiche2023vq} is crucial because it provides the tokens we use as training targets. Hence, the reconstruction error is transferred to the ground truth used to train MEGA. Fortunately, the reconstruction error of Mesh-VQ-VAE is about one order of magnitude smaller than the estimation error of SOTA HMR models. For predicting the rotation and camera parameters, we use the Euclidean distance on the rotation matrix corresponding to the predicted 6D representation and the L1 loss on the reprojection of 2D joints extracted from the predicted oriented mesh, using the predicted perspective camera parameters.

\begin{table*}[t]
    \centering
    \resizebox{.95\textwidth}{!}{
    \begin{tabular}{lc|ccc|ccc}
            \toprule
             & & \multicolumn{3}{c}{3DPW} & \multicolumn{3}{c}{EMDB} \\
            Method         & Backbone & PVE $\downarrow$   & MPJPE $\downarrow$ & PA-MPJPE $\downarrow$ & PVE $\downarrow$   & MPJPE $\downarrow$ & PA-MPJPE $\downarrow$       \\   \midrule
            FastMETRO~\cite{cho2022FastMETRO}    & HRNet-w64  & 121.6  & 109.0 & 65.7 & 119.2  & 108.1  & 72.7  \\
            PARE~\cite{Kocabas_PARE_2021}           & HRNet-w32 & 97.9   & 82.0  & 50.9 & 133.2  & 113.9  & 72.2  \\
            Virtual Marker~\cite{ma20233d} & HRNet-w48 & 93.8   & 80.5  & 48.9 & -  & - & - \\
            CLIFF~\cite{li2022cliff}          & HRNet-w48 & 87.6   & 73.9  & 46.4 & 122.9  & 103.1  & 68.8  \\
            VQ-HPS~\cite{fiche2023vq}         & HRNet-w48 & 84.8   & 71.1  & 45.2 & 112.9  & 99.9 & 65.2 \\
            MEGA (ours)  & HRNet-w48 & \textbf{81.6} & \textbf{68.5} & \textbf{44.1} & \textbf{107.9} & \textbf{90.5} & \textbf{58.7} \\
            \hspace{.5cm} \textit{linear masking}  & HRNet-w48 & 86.5 & 72.6 & 45.9 & 118.7 & 100.1 & 63.3 \\
            \hspace{.5cm} \textit{full mask} & HRNet-w48 & \underline{81.8} & \textbf{68.5} & \underline{44.4} & \underline{110.3} & \underline{92.7} & \underline{59.2} \\
            \hspace{.5cm} \textit{w/o pre-training + full mask}  & HRNet-w48 & 84.1 & \underline{70.5} & 46.2 & 113.9 & 95.9 & 62.0 \\ \midrule
            
            HMR2.0~\cite{goel2023humans}         & ViT-H & \underline{84.1} & \underline{70.0} & 44.5 & 120.1 & 97.8 & 61.5 \\
            TokenHMR$^\dagger$~\cite{dwivedi2024tokenhmr}  & ViT-H & 88.1 & 76.2 & 49.3 & 124.4 & 102.4 & 67.5  \\
            TokenHMR$^{\mp}$~\cite{dwivedi2024tokenhmr}  & ViT-H & 84.6 & 71.0 & \underline{44.3} & \underline{109.4} & \textbf{91.7} & \underline{55.6} \\
            MEGA (ours)  & ViT-H & \textbf{80.0} & \textbf{67.5} & \textbf{41.0} & \textbf{108.6} & \underline{92.4} & \textbf{52.5} \\
            \bottomrule
    \end{tabular}}
    \caption{\textbf{Evaluation in deterministic mode.} We evaluate MEGA on the 3DPW and EMDB datasets and compare it to the SOTA methods using metrics defined in \cref{subsec:setup} given in~mm. $\dagger$ stands for additionally using 2D training data, and $\mp$ for additionally using 2D data and BEDLAM~\cite{Black_CVPR_2023}. Methods in italic below the row "MEGA" indicate the results of the ablation study.}
    \label{tab:quant_det}
\end{table*}

\subsection{Generation strategy}\label{subsec:inference}

We propose two generation modes for inference:  deterministic and stochastic. Both modes start from a fully masked sequence of human mesh tokens,  aiming to generate a complete sequence that can be decoded into a canonical mesh. Regardless of the generation mode, the camera and rotation are predicted deterministically from the image.

\noindent{\bf Deterministic mode.} In deterministic inference mode, we predict all tokens in a single forward pass by taking the argmax of the predicted distribution over the human mesh token indices. In this mode, MEGA's encoder is not used; instead,  the decoder is fed with a sequence of $N$ mask tokens, relying entirely on the image representation information. Consequently,  the encoder can be discarded, significantly reducing the model size as $B_e > B_d$. To our knowledge, this is the first work that discards the encoder of an MAE, whereas previous works~\cite{he2022masked} typically discard the decoder and use the encoder to obtain representations for downstream tasks. 

\noindent{\bf Stochastic mode.} MEGA addresses the ambiguity of the HMR from a single image by operating in a stochastic inference mode and generating diverse plausible human meshes. We follow a strategy similar to~\cite{chang2022maskgit, chang2023muse, guo2023momask}, employing an iterative generation process in $T$ steps. At each step $t \in \{1,...,T\}$, we predict $n_t - n_{t-1}$ tokens, where $n_t = \lfloor N \times (1 - \cos(\frac{\pi t}{2T})) \rfloor$ denotes the number of predicted tokens up to step $t$. The tokens predicted at a given step remain visible for the subsequent steps. By the end of this iterative process, we generate $n_T = N$ tokens that represent a complete human mesh. The prediction at step $t \in \{1,...,T\}$ proceeds  as follows: 
\begin{itemize}[left=10pt]
\item The currently visible $n_{t-1}$ tokens are fed to the model.
\item For each of the $P_t = N-n_{t-1}$ tokens that are still masked, the model outputs unnormalized probabilities over the indices of the Mesh-VQ-VAE codebook. The corresponding categorical distributions are sampled using the Gumble-max trick~\cite{huijben2022review} to obtain $P_t$ candidate tokens to be set visible in the next step. Each candidate token is identified by an index between 1 and $S$ (the codebook size) along with its unnormalized probability.
\item Using the Gumble-max trick again, we finally sample $n_t - n_{t-1}$ tokens among the set of $P_t$ candidate tokens, which will be visible in the next step.
\end{itemize}

Given the stochastic nature of this generation mode, we can obtain $Q$ different human mesh predictions for a single image by repeating the above generation process several times. A visualization of the iterative generation process is shown in \cref{fig:hmr_process}.
\section{Experiments}\label{sec:experiments}

\subsection{Experimental setup}\label{subsec:setup}

\noindent{\bf Datasets.}
MEGA is initially pre-trained in a self-supervised manner on an extensive subset of the AMASS dataset~\cite{mahmood2019amass}, focusing on samples with high pose and body shape variety, as detailed in~\cite{Black_CVPR_2023}.
MEGA is then trained for  HMR  using a mix of standard image datasets labeled with pseudo-groundtruth human meshes~\cite{li2022cliff}, including MSCOCO~\cite{lin2014microsoft}, Human3.6M~\cite{ionescu2013human3}, MPI-INF-3DHP~\cite{mono-3dhp2017} and MPII~\cite{andriluka20142d}. We evaluate MEGA and compare it to the SOTA methods on the in-the-wild 3DPW~\cite{3dpw} and EMDB~\cite{kaufmann2023emdb} datasets. Following recent works~\cite{dwivedi2024tokenhmr, goel2023humans, fiche2023vq}, we do not finetune MEGA on the 3DPW training set before evaluation. This approach better validates the model's generalization capacity and allows us to use the same model for all experiments. Unless specified otherwise, other models in the comparison tables use the same training datasets and the same image feature extractor as ours for fair comparisons.  

\noindent{\bf Implementation details.} All  experiments are conducted  using PyTorch~\cite{paszke2019pytorch}. MEGA is pre-trained on human meshes for 500 epochs. For the HMR task,  MEGA is first trained on MSCOCO~\cite{lin2014microsoft} for 100 epochs, followed by training on the mix of datasets described above for 10 epochs. Using 4 NVIDIA A100 GPUs, the entire pre-training and training process takes about 2.5 days.

\noindent{\bf Metrics.} To evaluate HMR methods, we use the widely adopted metrics: per-vertex-error (PVE), mean-per-joint-error (MPJPE), and the Procrustes-aligned MPJPE (PA-MPJPE).   PVE measures the Euclidean distance between the vertices of the predicted mesh and the ground truth. MPJPE assesses the accuracy of body joints extracted from the mesh. PA-MPJPE is similar to MPJPE but includes a Procrustes-alignment, a rigid transformation that minimizes the distance between the predicted and ground truth joints. All three metrics are reported in~mm.

\subsection{Deterministic inference mode}

\noindent{\bf Quantitative evaluation.} In ~\cref{tab:quant_det}, we evaluate MEGA on the HMR task using the deterministic inference mode defined in~\cref{subsec:inference}, and compare its performance to  SOTA methods on 3DPW~\cite{3dpw} and EMDB~\cite{kaufmann2023emdb}. Results for SOTA methods are taken from the corresponding papers when available or computed using official implementations and model weights. To ensure fairness, we only compare methods trained on standard datasets (see~\cref{subsec:setup}), with the exception of TokenHMR~\cite{dwivedi2024tokenhmr}, which uses additional data but is included in the comparison because it also tokenized human pose. Note that the results for FastMETRO~\cite{cho2022FastMETRO} and Virtual Marker~\cite{ma20233d} differ from those in their original papers because we present results without finetuning on the 3DPW dataset. Similar to~\cite{fiche2023vq,dwivedi2024tokenhmr,goel2023humans}, we chose not to use the 3DPW training set in order to measure generalization. Furthermore, it allows us to use the same model to evaluate MEGA on the occlusion dataset 3DPW-OCC~\cite{zhang2020object} (see below) that contains sequences from the 3DPW training set. Qualitative results of MEGA in the deterministic inference mode on in-the-wild datasets~\cite{3dpw,kaufmann2023emdb} are shown in~\cref{app:additional}.

MEGA outperforms all other methods on in-the-wild datasets. Using an HRNet~\cite{wang2020deep} backbone, MEGA  significantly surpasses both parametric~\cite{Kocabas_PARE_2021,li2022cliff} and non-parametric~\cite{cho2022FastMETRO, ma20233d, fiche2023vq} methods, especially on the EMDB~\cite{kaufmann2023emdb} dataset. With a ViT~\cite{dosovitskiy2020image} backbone, MEGA  also achieves SOTA performance. However,  comparisons with TokenHMR~\cite{dwivedi2024tokenhmr} are not completely fair, as this method is trained with additional 2D data and BEDLAM~\cite{Black_CVPR_2023}. 

\noindent{\bf Ablation study.} \cref{tab:quant_det} also provides the results of an ablation study. We tried a different distribution for the variable masking rate, such that $M = \lfloor N \tau \rfloor$ with $\tau$ uniformly sampled in $[0,1[$ (``MEGA \textit{linear masking}'' in \cref{tab:quant_det}). Similar to~\cite{chang2022maskgit}, we find that the mask scheduling function improves the performance. This is also consistent with the findings of~\cite{he2022masked}, which showed that a masking rate higher than 50\% is beneficial for learning. We then tried masking 100\% of the human mesh tokens during the supervised training phase (``MEGA \textit{full mask}'' in \cref{tab:quant_det}). We could expect this ablation to perform better as the training procedure corresponds to the deterministic inference mode. However, the results are slightly degraded, suggesting that sharing the same procedure between the self-supervised and supervised training stages is important.  Additionally discarding the self-supervised pre-training of MEGA (``MEGA \textit{w/o pre-training + full mask}'' in \cref{tab:quant_det}) leads to a 2.5/6.0 mm increase in PVE on 3DPW/EMDB.

\noindent{\bf Results on an occlusion dataset} We quantitatively evaluate MEGA  on the occlusion dataset 3DPW-OCC~\cite{zhang2020object} in~\cref{tab:3dpw_occ} to evaluate the robustness of our approach to occlusions. Despite not being tailored for HMR on occluded images, MEGA surpasses all other methods trained on the same data (see~\cref{subsec:setup}). This performance could stem from MEGA's self-attention mechanism among mesh tokens.  While visible parts can be predicted by leveraging image embeddings, occluded parts heavily rely on visible body parts for accurate inference.

\begin{table}
    \centering
        \resizebox{.95\linewidth}{!}{
        \begin{tabular}{c|ccc}
            \toprule
            & \multicolumn{3}{c}{3DPW-OCC}\\
            Method  & PVE $\downarrow$   & MPJPE $\downarrow$ & PA-MPJPE $\downarrow$  \\ \midrule
            ROMP~\cite{sun2021monocular} & - & - & 65.9    \\
            SPIN~\cite{kolotouros2019learning}  & 121.6 & 95.6 & 60.8  \\
            VisDB~\cite{yao2022learning} & 110.5 & 87.3 & 56.0 \\
            3DCrowdNet~\cite{choi2022learning} & 103.2 & 88.6 & 56.8  \\
            SEFD~\cite{yang2023sefd} & \underline{97.1} & \underline{83.5} & \underline{55.0}  \\
            MEGA (ours) & \textbf{93.8} & \textbf{79.8} & \textbf{51.5} \\ \midrule
            PARE~\cite{Kocabas_PARE_2021} & 107.9 & 90.5 & 57.1  \\
            ScoreHypo~\cite{xu2024scorehypo} & \underline{89.8} & \underline{73.9} & \underline{48.7}  \\
            MEGA (ours) & \textbf{78.9} & \textbf{66.3} & \textbf{43.7} \\   
            \bottomrule
        \end{tabular}}
    \caption{\textbf{Evaluation on 3DPW-OCC.} We evaluate MEGA on an occlusion dataset and compare it to SOTA HMR methods designed to handle occlusions using standard metrics (see \cref{subsec:setup}) in~mm. Methods in the top part use a ResNet-50 backbone, while others use HRNet.}
    \label{tab:3dpw_occ}
\end{table}

\begin{table*}[t]
    \setlength{\tabcolsep}{4pt}
    \centering
    \resizebox{.95\textwidth}{!}{
    \begin{tabular}{c|ccccc|ccccc|ccccc}
            \toprule
            Method    & \multicolumn{4}{c}{PVE $\downarrow$} & Imp $\uparrow$ &  \multicolumn{4}{c}{MPJPE $\downarrow$} & Imp $\uparrow$  & \multicolumn{4}{c}{PA-MPJPE $\downarrow$} & Imp $\uparrow$   \\ 
            $Q$ & 1 & 5 & 10 & 25 &  & 1 & 5 & 10 & 25 &  & 1 & 5 & 10 & 25 &  \\ \midrule
            Diff-HMR~\cite{cho2023generative} & \underline{114.6} & \underline{111.8} & \underline{110.9} & \underline{109.8} & \underline{4.2} & 98.9 & 96.3 & 95.5 & 94.5 & 4.5 & \textbf{58.5} & 57.0 & 56.5 & 55.9 & 4.4 \\
            3D Multibodies~\cite{biggs2020} & - & - & - & - & - & \underline{93.8} & \underline{82.2} & \underline{79.4} & \underline{75.8} & \textbf{19.2} & 59.9 & 57.1 & 56.6 & 55.6 & 7.2 \\
            ProHMR~\cite{kolotouros2021probabilistic} & - & - & - & - & - & 97.0 & 93.1 & 89.8 & 84.0 & 13.4 & 59.8 & \underline{56.5} & \underline{54.6} & \underline{52.4} & \underline{12.4} \\
            MEGA (ours) & \textbf{101.6} & \textbf{92.8} & \textbf{90.4} & \textbf{87.5} & \textbf{13.9} & \textbf{86.2} & \textbf{78.0} & \textbf{76.4} & \textbf{73.9} & \underline{14.3} & \underline{58.6} & \textbf{51.6} & \textbf{49.7} & \textbf{47.6} & \textbf{18.7} \\
            MEGA det (ours) & \multicolumn{5}{l}{90.6} & \multicolumn{5}{|l}{76.3} & \multicolumn{5}{|l}{48.3} \\ \midrule
            
            ScoreHypo$^\ddagger$~\cite{xu2024scorehypo} & - & - & - & - & - & - & \underline{75.3} & \underline{71.7} & \underline{67.8} & - & - & \underline{47.4} & \underline{45.2} & \underline{42.5} & - \\
            MEGA$^\ddagger$ (ours) & \textbf{83.4} & \textbf{78.4} & \textbf{76.9} & \textbf{75.1} & \textbf{10.0} & \textbf{69.9} & \textbf{66.1} & \textbf{64.9} & \textbf{63.6} & \textbf{9.0} & \textbf{45.5} & \textbf{42.6} & \textbf{41.7} & \textbf{40.4} & \textbf{11.2} \\
            MEGA det$^\ddagger$ (ours) & \multicolumn{5}{l}{81.6} & \multicolumn{5}{|l}{68.5} & \multicolumn{5}{|l}{44.1} \\
            \bottomrule
        \end{tabular}}
    \caption{\textbf{Evaluation in stochastic mode.} We compare MEGA to the SOTA probabilistic methods on the multi-output HMR task using standard metrics (see \cref{subsec:setup}) given in~mm and the relative improvement (Imp) in \%. $\ddagger$  uses an HRNet backbone; all other methods use a ResNet-50 backbone.}
    \label{tab:quant_stoch}
\end{table*}

\begin{figure}
    \centering
    \includegraphics[width=.9\linewidth]{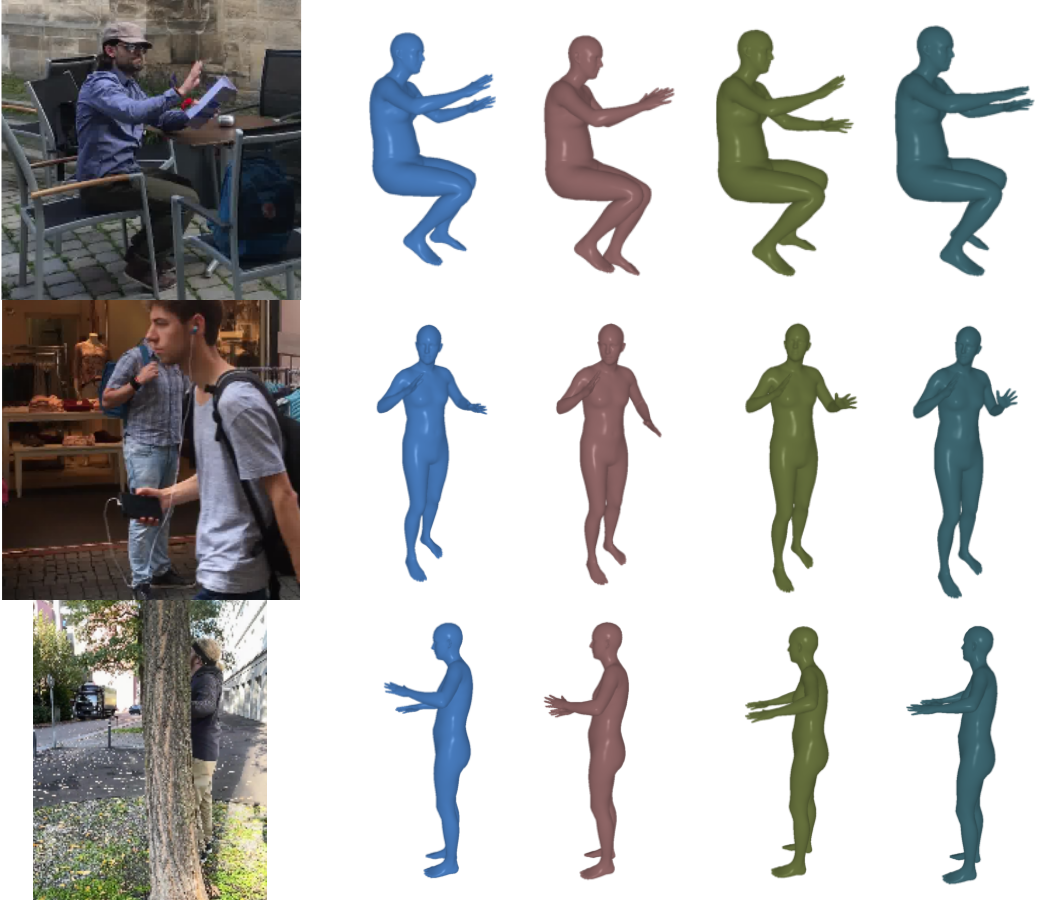}
    \caption{\textbf{Qualitative samples.} Given a single image with occlusions, MEGA makes diverse plausible predictions.}
    \label{fig:qual_stoch}
\end{figure}

\subsection{Stochastic inference mode}

\noindent{\bf Quantitative evaluation.} We evaluate MEGA in stochastic inference mode (see~\cref{subsec:inference}) for the HMR task on the 3DPW~\cite{3dpw} dataset, comparing our performance with  SOTA multi-output approaches. Results from other methods are obtained from their respective papers. We exclusively compare methods trained on standard datasets (see~\cref{subsec:setup}), providing a version of MEGA using a ResNet-50~\cite{he2016deep} backbone for fair comparisons. Following prior works~\cite{cho2023generative,biggs2020,kolotouros2021probabilistic}, we assess accuracy using standard metrics computed with the minimum error sample out of $Q$ predictions, with $Q$ ranging from $1$ to $25$. We also calculate the relative improvement between 1 and 25 samples. Quantitative results are presented in \cref{tab:quant_stoch} and qualitative samples are shown in \cref{fig:qual_stoch}.

MEGA achieves SOTA performance across all metrics and sample sizes. Notably, MEGA exhibits significantly higher accuracy than other methods when generating a single sample, and consistently demonstrates the best or second-best relative improvement among methods using a ResNet-50 backbone.  With an HRNet~\cite{wang2020deep} backbone, MEGA  outperforms all probabilistic methods and surpasses SOTA single-output methods, even with a single stochastic generation.  It is worth noting that it takes around 10 stochastic samples to outperform the deterministic generation (MEGA det), highlighting the utility of deterministic mode for quick and accurate predictions and the advantage of multiple predictions for enhanced accuracy. For a detailed comparison between deterministic and stochastic modes and a discussion on interpreting diverse predictions, please refer to~\cref{app:error_plot} and~\cref{app:interpretation}.

\noindent{\bf Generation process.} We provide visualizations of the prediction process in stochastic mode.  Specifically, using the model with an HRNet backbone that generates meshes in 5 steps (see \cref{app:implementation_details}), given an image, we visualize all intermediate steps. Results are shown in~\cref{fig:hmr_process}. Importantly, our model does not predict unrealistic meshes in the first steps (all masked tokens were replaced by the index 0 for visualization purposes). We observe that a rough estimate of the mesh is provided even in the initial steps, where only a few indices are predicted (2 during the first iteration and 8 during the second as shown in \cref{fig:hmr_process}). This outcome is expected because the tokens set to be visible after each step are the most likely. Consequently, the indices with the highest confidence are selected first, enabling the construction of a preliminary mesh estimate. Subsequent steps refine these predictions, enhancing their realism.

\section{Conclusion}\label{sec:conclusion}
In this work, we explored self-supervised learning and masked generative modeling on human meshes for HMR. We introduced MEGA, a masked generative autoencoder designed to generate human meshes as discrete token sequences. MEGA's flexible architecture and generation scheme enables the generation of diverse and realistic meshes and supports both single and multi-output HMR. An extensive evaluation demonstrates significant improvement over SOTA in both these domains. Further discussions on the limitations and future works are available in \cref{app:discussion}.

\section*{Acknowledgment}
This study is part of the EUR DIGISPORT project supported by the ANR within the framework of the PIA France 2030 (ANR-18-EURE-0022). This work was performed using HPC resources from the “Mésocentre” computing center of CentraleSupélec, ENS Paris-Saclay, and Université Paris-Saclay supported by CNRS and Région Île-de-France. We would like to thank Samir Sadok for his valuable help with the implementation of the masked autoencoder.
{
    \small
    \bibliographystyle{ieeenat_fullname}
    \bibliography{main}
}

\clearpage
\setcounter{page}{1}
\setcounter{section}{0}
\renewcommand{\thesection}{\Alph{section}}
\maketitlesupplementary

\begin{figure*}
    \centering
    \includegraphics[trim={0 18cm 0 20cm},clip, width=\textwidth]{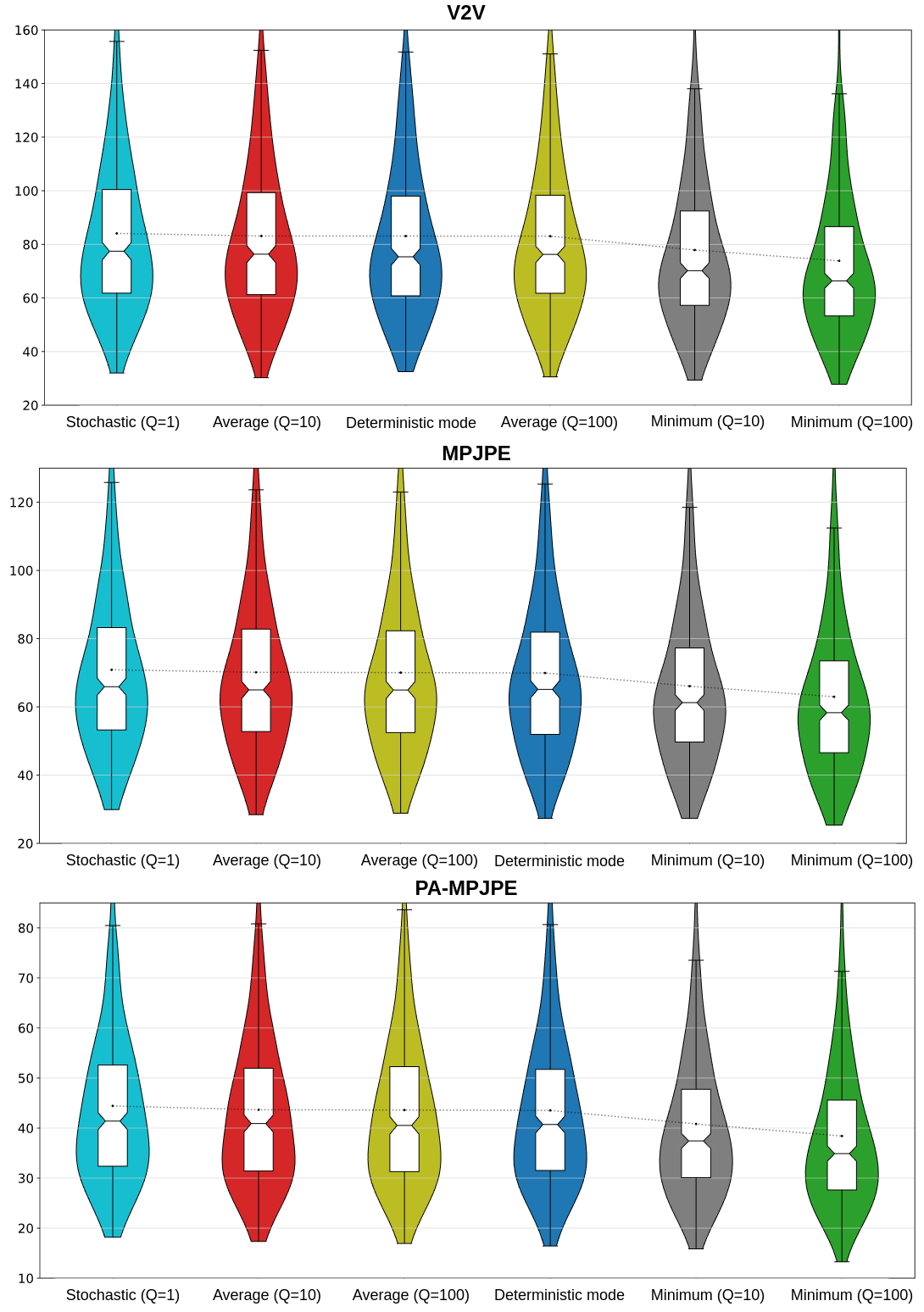}
    \caption{\textbf{Error distribution.} We visualize the distribution of the MPJPE in~mm on the 3DPW dataset. }
    \label{fig:plot_err}
\end{figure*}

\section{Link between the deterministic and stochastic modes}\label[appendix]{app:error_plot}

\begin{table}
    \centering
    \resizebox{\linewidth}{!}{
    \begin{tabular}{lccccccc}
    \toprule
    & \multicolumn{6}{c}{Stochastic} & Deterministic \\
    $Q$ & 1 & 5 & 10 & 25 & 50 & 100 & \\ \midrule
    PVE $\downarrow$ & 84.08 & 83.27 & 83.13 & 83.09 & 83.02 & 83.05 & 83.10 \\
    MPJPE $\downarrow$ & 70.88 & 70.29 & 70.15 & 70.09 & 70.04 & 70.05 & 69.95 \\
    PA-MPJPE $\downarrow$ & 44.42 & 43.77 & 43.66 & 43.65 & 43.62 & 43.61 & 43.53 \\
    Dist. to det. & 14.78 & 9.53 & 8.56 & 7.95 & 7.71 & 7.61 & 0.00 \\
    SD & - & 11.61 & 12.30 & 12.65 & 12.79 & 12.88 & - \\
    \bottomrule
    \end{tabular}}
    \label{tab:det_vs_stoch}
    \caption{\textbf{Comparison between deterministic and stochastic generation modes.} In stochastic mode, we evaluate the mean mesh obtained with different sample sizes on 10\% of the 3DPW~\cite{3dpw} dataset, and we provide its distance to the deterministic prediction (Dist. to det.). We also report the standard deviation of the predictions. All metrics are in~mm.}
\end{table}

To gain deeper insights into the stochastic generation mode, we propose not only evaluating the best sample among the $Q$ generations (the common practice in the literature), but also assessing the mean of the generated meshes. In~\cref{tab:det_vs_stoch}, we compare the performance of the average prediction for different $Q$ using the standard metrics (refer to~\cref{subsec:setup}). Additionally, we compute the Euclidean distance between the mean mesh and the one obtained in deterministic mode (Dist. to det., in mm) and the standard deviation of the predictions averaged over all the vertices (SD, in~mm). To reduce the computational costs, this study is conducted on a randomly selected 10\% subset of images from the 3DPW dataset. \cref{tab:det_vs_stoch} shows that as $Q$ increases, the average prediction in stochastic mode approaches the deterministic prediction, with a distance around 7~mm for $Q=100$. The average prediction appears to converge toward a favorable solution,   slightly outperforming the deterministic prediction in terms of PVE.

We also examine the distributions of the MPJPE on the 3DPW~\cite{3dpw} dataset for MEGA in both deterministic and stochastic modes across various sample sizes $Q$. In stochastic mode, we analyze the average and best predictions. The results are reported in~\cref{fig:plot_err}. Notably, the average prediction error of the stochastic mode appears to converge toward the deterministic prediction error, particularly as $Q$ increases; the distributions are very similar, with overlapping 95\% confidence intervals. When $Q$ equals 1, the mean performance is comparatively lower, resulting in higher error values. These observations underscore the importance of having a deterministic mode for rapid and accurate predictions, which can be considered an estimator of the average prediction over $Q$ samples.

When selecting the prediction with the minimum error among $N$ samples, we observe a shift in the distribution shapes, with errors concentrated toward lower values.   While the highest error values decrease notably, the lowest remain relatively unchanged.  This phenomenon likely occurs because the lowest errors typically correspond to meshes that are easier to predict and exhibit lower standard deviation.  Consequently, the stochastic mode proves particularly beneficial for challenging images, where multiple predictions offer valuable insights.

\section{Experimental details}\label[appendix]{app:implementation_details}

\noindent{\bf Pre-training stage.} As detailed in the main body, the pre-training stage is done on a subset of AMASS~\cite{mahmood2019amass}, as introduced in~\cite{Black_CVPR_2023}. We pre-train MEGA for 500 epochs, a task accomplished in less than a day on 4 A100 GPUs.  MEGA is trained using the AdamW optimizer~\cite{loshchilov2017decoupled} with a cosine scheduler to adjust the learning rate. The base learning rate is $1e-3$, and we have a warmup of 20 epochs. The optimizer's parameters are $\beta_1=0.9$, $\beta_2=0.99$, and the weight decay is 0.05.

\noindent{\bf Supervised training stage.} The supervised training stage is done on a mix of standard datasets for HMR as presented in~\cref{subsec:setup}. We first train MEGA on MSCOCO~\cite{lin2014microsoft} for 100 epochs and then train on the whole training set for 10 epochs. Each step takes about 1 day on 4 A100 GPUs. The training settings are exactly the same as the pre-training regarding learning rates and schedulers. For each training step, we start at epoch 0. Note that we have a lower learning rate in practice for the training on the mix of datasets because we stop the training before finishing the warm-up period. For training with HRNet and ResNet-50 backbones, the weights of the backbone are fine-tuned with the same settings as the other parameters of MEGA. When using ViT, the backbone is frozen during the training on MSCOCO, and we only fine-tune the last 10 blocks when training on standard datasets for computing power reasons.

\noindent{\bf HMR.} For recovering human meshes from images in deterministic mode, we predict all images in a single step without randomness. In stochastic mode, we have to set the number of steps for generating the sequence of human mesh tokens and the amount of noise injected for the Gumble-max sampling. Note that we did not test MEGA with a ViT in stochastic mode. With HRNet, we generate meshes in 5 steps, and the initial noise temperature is 1. The generation process with ResNet-50 is made in 2 steps, and the initial noise temperature is fixed to 10. The amount of noise at step $t$ is $A \times (1 - \frac{t}{T})$ where $A$ is the initial noise temperature~\cite{chang2023muse,chang2022maskgit}.

\noindent{\bf Random meshes generation.} We generate random meshes in 20 steps (see next section for more details). We want the generation to be completely random for the first steps so that the predictions are diverse. However, the last steps should be almost deterministic to obtain realistic meshes. The initial noise temperature is $A=1.2$, and the amount of noise at step $t$ is given by $(A \times (1 - \frac{t}{T}))^6$.

\noindent{\bf Inference time} In deterministic mode, with a batch size of 1, a forward pass takes 0.07 seconds with the HRNet version of MEGA on a GeForce GTX 1070 GPU. With similar settings, the ResNet version is real-time (0.03 seconds). In the stochastic mode, generating a single prediction with the ResNet model takes 0.04 seconds, which is still real-time, and generating 16 predictions takes 0.23 seconds.

\begin{figure*}
\centering
\includegraphics[trim=25 5 20 2, clip, width=\textwidth]{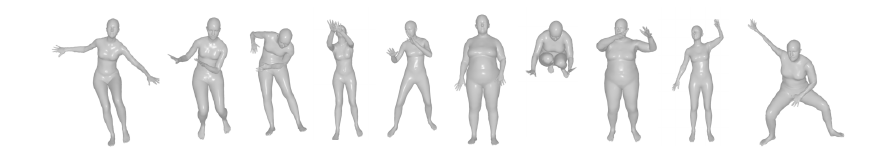}
\caption{\textbf{Random mesh generations.} We use MEGA pre-trained in a self-supervised fashion to generate random human meshes.}
\label{fig:rand_gen}
\end{figure*}

\section{Random meshes generation}\label[appendix]{app:rand_gen}

We propose to use MEGA pre-trained in a self-supervised manner (see~\cref{subsec:train}) for generating random human meshes. For comparison, we assess its capabilities against VPoser~\cite{pavlakos2019expressive} and NRDF~\cite{he2024nrdf}. VPoser is a conventional pose prior in  VAE form, while NRDF is a SOTA pose prior based on neural fields~\cite{xie2022neural}. Although these models do not explicitly model body shapes, making them not directly comparable to MEGA, which directly generates meshes with diverse poses and shapes, they are the most suitable for comparison purposes. As far as we know, MEGA is the first model generating unconditioned random human body meshes with pose and shape diversity. We assess all 3 models in terms of diversity using the average pairwise distance (APD) in~cm, representing the average distance between the joints of all pairs of samples. For plausibility evaluation, we compute the Fréchet inception distance (FID) with the fully convolutional mesh autoencoder introduced in~\cite{zhou2020fully} trained on AMASS~\cite{mahmood2019amass}, with a latent space dimension of $7 \times 9$. The FID compares the latent representation of generated meshes with that of a representative subset of AMASS introduced in~\cite{Black_CVPR_2023}.

We randomly sample 500 meshes with each method. Regarding plausibility, MEGA outperforms other methods, achieving an FID of 0.001 compared to 0.007 for VPoser and 0.033 for NRDF. This result is not surprising, as other methods use the average shape for all meshes, whereas MEGA produces diverse results in poses and shapes. NRDF generates more diverse meshes, with an APD of 28.61~cm, while VPoser and MEGA achieve APDs of 18.32 and 20.77~cm, respectively. In summary, MEGA clearly outperforms VPoser, as our generated samples are more diverse and plausible. NRDF produces more diverse poses, but the distribution of the generation samples of MEGA is more representative of the AMASS dataset. In~\cref{fig:rand_gen}, we present some qualitative samples of MEGA's generation, which exhibit diverse and realistic poses and shapes.

\begin{figure*}[t!]
    \centering
    \includegraphics[width=\textwidth]{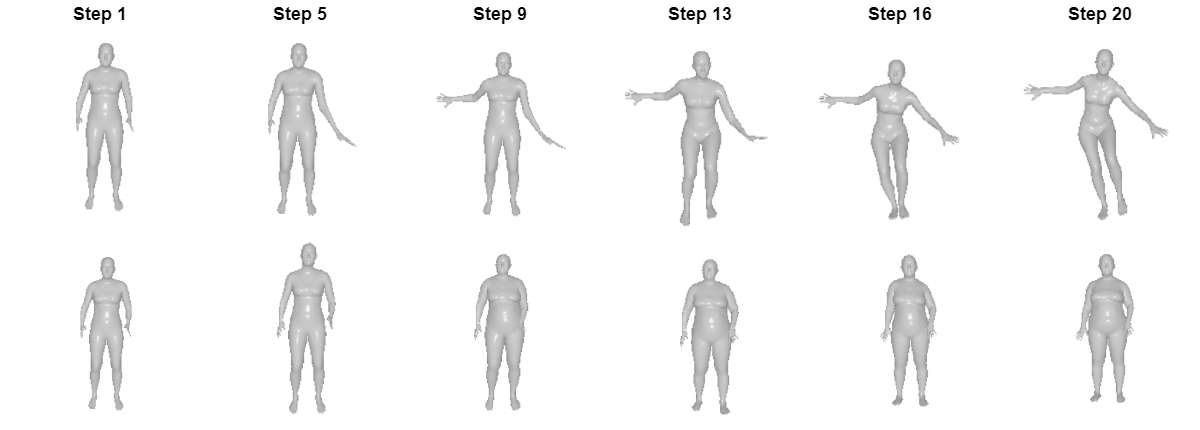}
    \caption{\textbf{Prediction process iterations.} We visualize the predictions for intermediate random generations. All masked tokens are replaced by the first token of the codebook, corresponding to index 0.}
    \label{fig:gen_process}
\end{figure*}

In \cref{fig:gen_process}, similar to the main paper, we visualize the generation process after 1, 5, 9, 13, 16, and 20 steps. The initial mesh appears almost identical across all generations, as only one token is predicted in the first step (with all others set to 0 for visualization). However, diversity quickly emerges in subsequent steps, with the final steps refining the meshes to be more realistic. This pattern was anticipated, as the initial steps involve considerable randomness, whereas the later steps tend to become more deterministic.

\begin{figure*}
    \centering
    \includegraphics[width=0.9\textwidth]{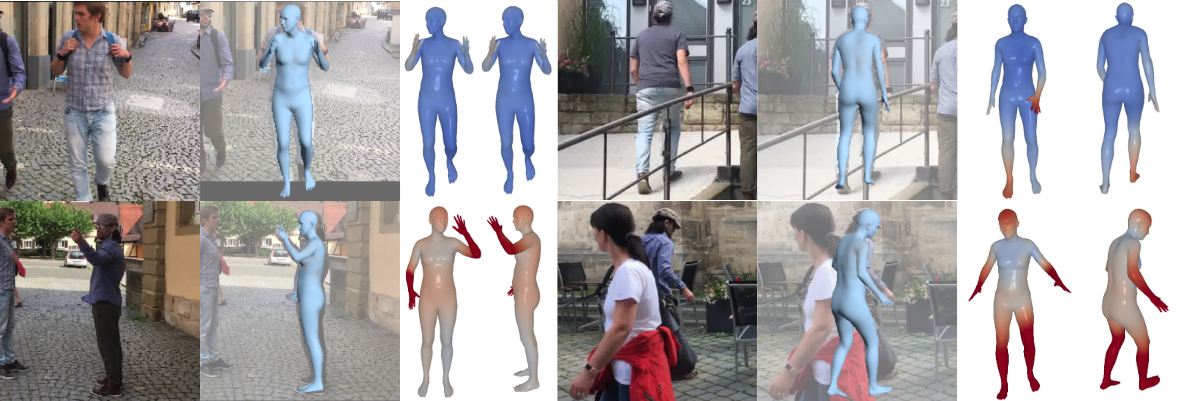}
    \caption{\textbf{Visualization of the predictions diversity.} We visualize the standard deviation of the 3D location of each vertex. {\textcolor{blue}{Bluish}} regions in the mesh indicate low standard deviation, while {\textcolor{red}{reddish}} areas signify higher standard deviation.}
    \label{fig:uncertainty}
\end{figure*}

\section{Interpreting diversity in predictions}\label{app:interpretation}

In the stochastic mode, MEGA makes diverse predictions given a single image. After making multiple predictions given an image, we can compute the standard deviation of the position of each vertex and interpret this value as a measure of the uncertainty in predictions. Indeed, if all samples are similar, we can conclude that the model is ``certain" about this prediction. When the results are very diverse given a single image, we can interpret that as high uncertainty.

Some qualitative samples are shown in \cref{fig:uncertainty}. The two images on the top show non-occluded bodies, and the relative depth between body parts is easy to perceive. Thus, the standard deviation is low; the model consistently makes accurate predictions. The body is partially occluded in the two bottom images, and the depth of some body parts (such as the left arm in the left image) is hard to estimate. The model makes diverse predictions, which can be interpreted as high uncertainty.

\begin{figure*} [t!]
    \centering
    \includegraphics[width=0.8\textwidth]{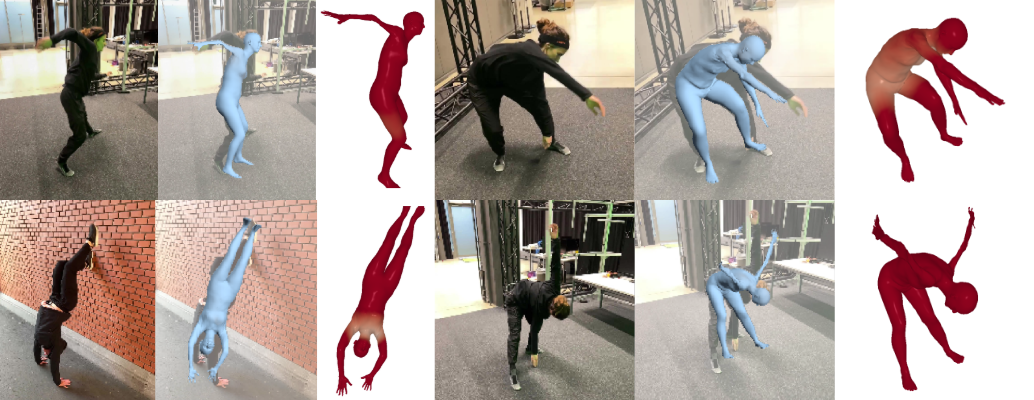}
    \caption{\textbf{Failure cases.} In failure cases, it is worth noting that our model predicts very diverse results, which can be interpreted as high uncertainty.}
    \label{fig:failures}
\end{figure*}

\section{Qualitative results and failure cases}\label[appendix]{app:additional}

We present several failure cases in~\cref{fig:failures}. Extreme poses can result in prediction errors, occasionally leading to non-anthropomorphic predictions as we do not rely on the SMPL parameters~\cite{loper2015smpl}. Notably, the standard deviation of the vertices' position is exceptionally high in such instances.

\begin{figure*} [t!]
    \centering
    \includegraphics[width=\textwidth]{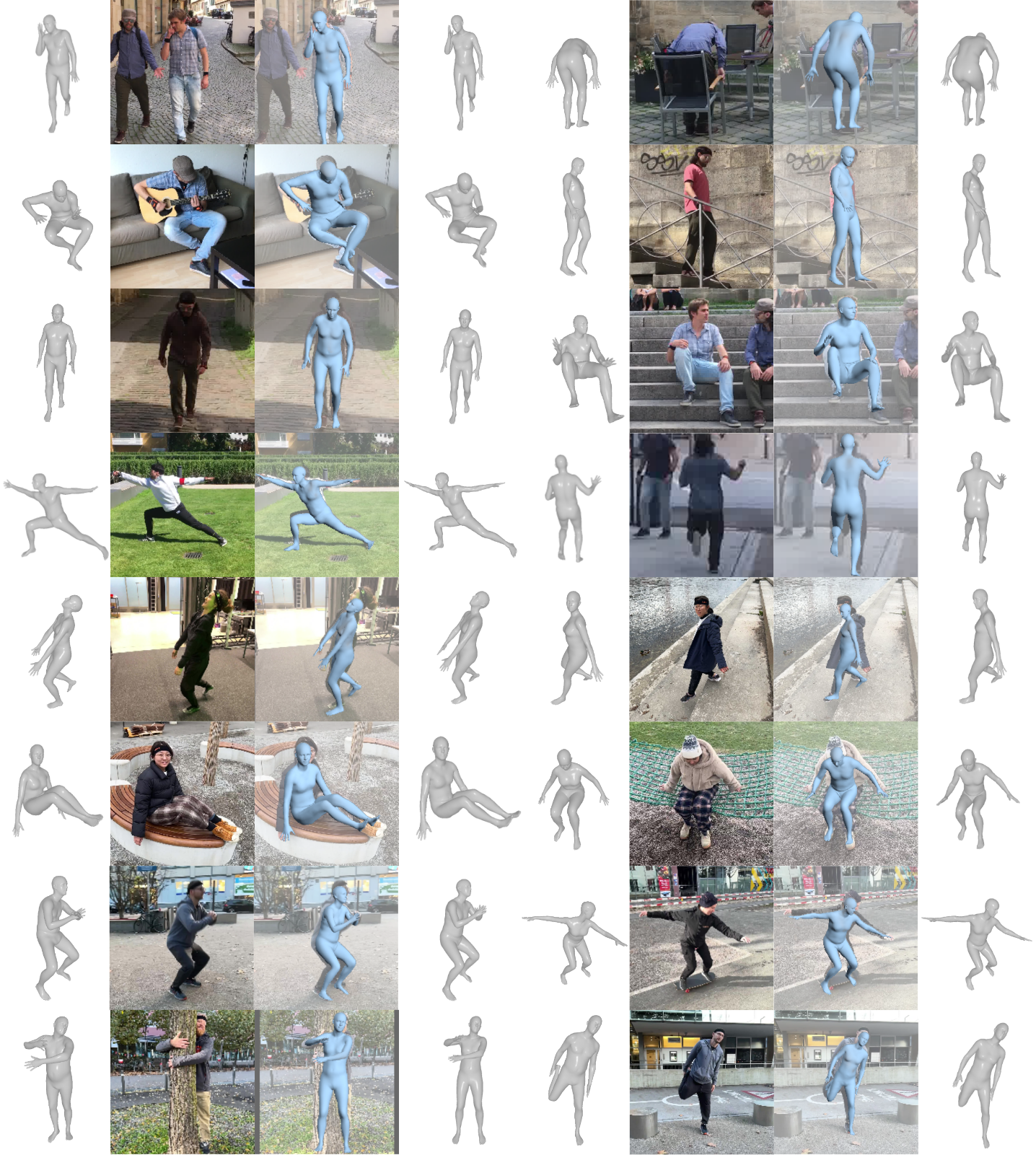}
    \caption{\textbf{Qualitative examples on 3DPW and EMDB.} The first column is the groundtruth, the second and third are the image and the reprojection of the deterministic prediction, and the fourth is the prediction using the deterministic mode of MEGA with an HRNet backbone.}
    \label{fig:qualitative}
\end{figure*}

\cref{fig:qualitative} presents qualitative samples from in-the-wild datasets.  We can observe that in some cases (for instance, images in the third and fourth rows on the right), our predictions appear even more accurate than the ground truth. While this result is encouraging, it underscores the limited value of striving for fractions of millimeters of accuracy on datasets like 3DPW when the ground truth itself is imperfect.

\section{Further discussions}\label{app:discussion}

\noindent{\bf Tokenizing vertices vs. SMPL parameters.} Predicting the SMPL parameters presents weaknesses~\cite{fiche2023vq}, such as error accumulation in the kinematic tree, that are addressed when working on the 3D vertices. The SMPL model parameters are attractive because they are a dense representation of human meshes, but once tokenized, this advantage is no longer leveraged. Even if the tokenization of Mesh-VQ-VAE starts from high-dimensional data, the obtained discrete representation is much denser than the representation proposed in TokenHMR in terms of sequence length ($54$ vs. $160$) and codebook size ($512 \times 8$ vs. $2048 \times 256$): it is easier to predict 54 tokens among 512 values than 160 tokens with 2048 possible values.

\noindent{\bf Limitations.} While MEGA generally produces accurate predictions, it struggles with extreme poses significantly divergent from the training data. \cref{fig:failures} in \cref{app:additional} provides visualizations of these failure cases. 

\noindent{\bf Future work.} MEGA's adaptable framework suggests potential applications beyond its current scope. Future research could explore generating human meshes conditioned on text inputs~\cite{delmas2022posescript}. We could also complement image embedding with more observations, such as 2D pose tokens~\cite{geng2023human} or tokenized meshes of other individuals to model social interactions~\cite{guo2022multi, muller2023generative}. Extending this work to videos by incorporating temporal masking during training~\cite{sadok2023avector,sadok2023vector} or including more extreme poses in training data~\cite{tripathi20233d} may improve performance. Future works may also focus on other generative models such as discrete diffusion~\cite{austin2021structured}.

\noindent{\bf Potential applications.} A direct application of MEGA is to generate solutions until we have a satisfactory answer, similar to LLMs. ScoreHypo [87] proposed a scoring network to select the best prediction among a range of outputs, increasing the accuracy of predictions. We can also choose the most suitable prediction depending on the use case: the solution that minimizes the re-projection error in sports applications for higher precision, the most visually appealing result in animation... The diversity of predictions can also be interpreted as a measure of per-vertex uncertainty~\cite{le2024meshpose} (see~\cref{app:interpretation}).

\noindent{\bf Broader impact.}  MEGA contributes to the understanding of human perception from images. While there is concern about potential misuse for intrusive surveillance, MEGA does not reconstruct facial features, preserving anonymity. MEGA could have applications in healthcare, such as motor assessment of patients. This application would be positive, but potential prediction errors could negatively affect the care pathway. 

\end{document}